\documentclass{Interspeech2024}

\interspeechcameraready

\usepackage{float}
\usepackage{booktabs}
\usepackage{multirow}
\usepackage{multicol}
\newcommand{\p}{\mathrm{p}}
\usepackage{cite}
\usepackage{hyperref}
\hypersetup{
    colorlinks=true,
    linkcolor=red,
    urlcolor=magenta,
    }
    
\title{Confidence Estimation for Automatic Detection of Depression and\\ Alzheimer's Disease Based on Clinical Interviews}

\name[affiliation={1}]{Wen}{Wu}
\name[affiliation={2}]{Chao}{Zhang}
\name[affiliation={1}]{Philip C.}{Woodland}

\address{
  $^1$Department of Engineering, University of Cambridge, Cambridge, UK\\
  $^2$Department of Electronic Engineering, Tsinghua University, Beijing, China\thanks{W. W. is supported by a Cambridge International Scholarship from the Cambridge Trust.} }
\email{\{ww368, pw117\}@cam.ac.uk, cz277@tsinghua.edu.cn}


\begin{document}

\maketitle

\begin{abstract}
 Speech-based automatic detection of Alzheimer's disease (AD) and depression has attracted increased attention. Confidence estimation is crucial for a trust-worthy automatic diagnostic system which informs the clinician about the confidence of model predictions and helps reduce the risk of misdiagnosis. This paper investigates confidence estimation for automatic detection of AD and depression based on clinical interviews. A novel Bayesian approach is proposed which uses a dynamic Dirichlet prior distribution to model the second-order probability of the predictive distribution. Experimental results on the publicly available ADReSS and DAIC-WOZ datasets demonstrate that the proposed method outperforms a range of baselines for both classification accuracy and confidence estimation.

\end{abstract}

\section{Introduction}

In recent years, awareness and concern for mental health, such as Alzheimer's disease (AD) and depression, have significantly increased. AD is a progressive neurodegenerative disorder marked by diminishing cognitive abilities, memory loss, and a decline in everyday skills. It is the leading cause of dementia, impacting millions across the world \cite{mattson2004pathways}. Meanwhile, depression is a widespread mental disorder that manifests through ongoing sadness, a lack of interest in activities previously enjoyed, and also diminished thinking capabilities. Depression affects all age groups and cultural backgrounds, potentially causing persistent thoughts of death or suicidal ideation \cite{edition2013diagnostic}.

The precise diagnosis of AD and depression is essential for their effective management and the initiation of timely intervention. This has driven forward research in the automatic detection of AD~\cite{ivanov2013phonetic,mirheidari2018detecting,cui2023transferring} and depression~\cite{moore2007critical,yu2015cognitive,he2021automatic,wu2023self}. Studies have explored a variety of hand-crafted features, including acoustic aspects like pitch variation, syllable rate, and spectrogram analysis~\cite{ivanov2013phonetic,mirheidari2018detecting,moore2007critical,yu2015cognitive,low2010detection,ooi2012multichannel}, along with linguistic aspects such as part-of-speech information, sentence structure, and vocabulary diversity~\cite{yang2016decision,gong2017topic,bucks2000analysis,fraser2016linguistic}. The advent of deep learning pre-trained speech and language models, such as WavLM~\cite{chen2022wavlm}, Whisper~\cite{radford2023robust}, and BERT~\cite{devlin-etal-2019-bert}, has offered promising results for extracting features relevant to diagnosing AD and depression~\cite{cui2023transferring,wu2023self,balagopalan2020bert,yuan2020disfluencies,syed2020automated,wu2022climate}.

Estimating confidence levels is key in medical tasks. While supportive in diagnosis, deep learning models often suffer from calibration issues, leading to high confidence in incorrect predictions (\textit{i.e.} confidently wrong). Confidence estimation can increase the reliability and interpretability of diagnostics powered by deep learning, offering clinicians a clearer understanding of how much trust to place in automated predictions to reduce the risk of misdiagnosis. It can also facilitate the identification of ambiguous and borderline cases, necessitating the input of clinical expertise. While confidence estimation techniques have been applied in areas like speech recognition~\cite{wessel2001confidence,jiang2005confidence,yu2011calibration,li2021confidence} and dialogue systems~\cite{tur2005combining}, their application in detecting mental illnesses through speech analysis remains largely unexplored.

This paper investigates confidence estimation for automatic AD and depression detection based on speech recordings from clinical interviews. Standard deep neural network classifiers are often trained to maximise the categorical probability of the correct class using the cross-entropy loss, whose output logit values are converted into pseudo-categorical distributions as the predictions using a softmax function. 
This standard framework is known to have unreliable uncertainty estimations~\cite{guo2017calibration,gal2016dropout}. In this paper, a novel Bayesian approach is proposed which introduces a dynamic Dirichlet prior distribution to model the second-order probability over the predictive distribution. The dynamic Dirichlet prior is predicted by a deep neural network trained by minimising the Bayes risk of the prediction. Multiple evaluation metrics are adopted to evaluate the proposed method in terms of classification performance and confidence estimation. Results on the ADReSS and DAIC-WOZ datasets show that the proposed method outperforms the baselines in terms of both classification accuracy and model calibration.
To the best of our knowledge, this is the first work that investigates confidence estimation for automatic AD and depression detection based on clinical interviews.

The rest of the paper is organised as follows. Section~\ref{sec: method} introduces the proposed approach of confidence estimation. Evaluation metrics and experimental setup are presented in Sections~\ref{sec: metric} and~\ref{sec: exp setup} respectively. Experimental results are shown in Section~\ref{sec: results}, followed by the conclusions.

\section{Confidence estimation method}
\label{sec: method}
Confidence is defined as the probability corresponding to the predicted class in the predictive distribution.
A standard neural network classifier predicts a categorical distribution which represents the probabilistic assignment over the possible classes.
Inspired by evidential deep learning~\cite{sensoy2018evidential,malinin2018predictive, wu2022estimating}, which was originally designed for detection of out-of-distribution samples, we introduce a novel Bayesian approach for confidence estimation which places a dynamic Dirichlet prior over the categorical distribution to measure the probability of the predictive distribution (\textit{i.e.,} second-order probability) instead of a point estimate.

\subsection{Modelling second-order probability}
Consider the target label as a one-hot vector $\boldsymbol{t}$ where $t_k$ is one if class $k$ is the correct class else zero. $\boldsymbol{t}$ is sampled from a categorical distribution $\boldsymbol{\pi}$ where each component ${\pi}_k$ corresponds to the probability of sampling a label from class $k$:
\begin{equation}
    \boldsymbol{t} \sim \mathrm{P}(\boldsymbol{t}|\boldsymbol{\pi}) = \operatorname{Cat}(\boldsymbol{t} | \boldsymbol{\pi}) = {\pi}_k^{t_k}.
    \label{eqn: cat}
\end{equation}
A Dirichlet prior is introduced to model the distribution over the categorical distribution. Assume the categorical distribution is drawn from a Dirichlet distribution:
\begin{equation}
    \boldsymbol{\pi} \sim \mathrm{p}(\boldsymbol{\pi} | \boldsymbol{\alpha}) = \operatorname{Dir}(\boldsymbol{\pi} | \boldsymbol{\alpha}) = \frac{1}{B(\boldsymbol{\alpha})}\prod_{k=1}^K \pi_k^{\alpha_k-1}
\end{equation}
where $\boldsymbol{\alpha}$ is the hyperparameter of the Dirichlet distribution and $K$ is the number of classes. $ B(\cdot)$ is the Beta function:
\begin{equation}
    \mathrm{B}(\boldsymbol{\alpha})=\frac{\prod_{k=1}^K \Gamma\left(\alpha_k\right)}{\Gamma\left(\sum_{k=1}^K \alpha_k\right)}
\end{equation}
where $\Gamma(\cdot)$ denotes the Gamma function $\Gamma(z)=\int_0^\infty x^{z-1} e^{-x} dx$.

\begin{figure}[t]
    \centering
    \includegraphics[width=0.8\linewidth]{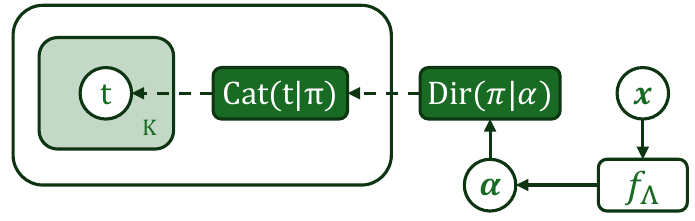}
    \caption{Illustration of the modelling process. 
    }
    \label{fig: illus}
\end{figure}

The output of a standard neural network classifier is a probability assignment over the possible classes. The Dirichlet distribution represents the density of each such probability assignment, hence modelling second-order probabilities and uncertainty. 
For a given input $\boldsymbol{x}^{(i)}$, the hyperparameter $\boldsymbol{\alpha}^{(i)}$ is predicted by a neural network $\boldsymbol{\alpha}^{(i)}=\boldsymbol{f_\Lambda} (\boldsymbol{x}^{(i)})$ where $\boldsymbol{\Lambda}$ is the model parameters. An exponential activation is applied to the model output to ensure $\boldsymbol{\alpha}^{(i)}$ is positive. The modelling process is illustrated in Figure~\ref{fig: illus}.

\subsection{Learning a dynamic Dirichlet prior}
Given a one-hot label $\boldsymbol{t}^{(i)}$ and predicted Dirichlet $\operatorname{Dir}(\boldsymbol{\pi}^{(i)} | \boldsymbol{\alpha}^{(i)})$, a neural network can be trained by minimising the Bayes risk with respect to the sum of squares loss between targets and the class predictor:
\begin{equation}
    \begin{aligned}
&\mathcal{L}^{(i)}_{\text{BR}}(\boldsymbol{\Lambda}) =\int \parallel \boldsymbol{t}^{(i)}-\boldsymbol{{\pi}}^{(i)} \parallel ^2 \p(\boldsymbol{\pi}^{(i)}|\boldsymbol{\alpha}^{(i)}) d\boldsymbol{\pi}^{(i)} \\
& =\sum_{k=1}^K \mathbb{E}\left[(t^{(i)}_{k})^2-2 t^{(i)}_{k} \pi^{(i)}_{k}+(\pi^{(i)}_{k})^2\right]\\
& =\sum_{k=1}^K \left(t_{k}^{(i)}- \frac{\alpha_{k}^{(i)}}{\alpha_0^{(i)}} \right)^2+\frac{\alpha_k^{(i)}\left(\alpha_0^{(i)}-\alpha_k^{(i)}\right)}{(\alpha_0^{(i)})^2\left(\alpha_0^{(i)}+1\right)}.
\end{aligned}
\end{equation}
Following~\cite{wu2022estimating}, KL divergence between targets and prediction is introduced as an additional regularisation term.
The total loss is then defined as:
\begin{equation}
    \mathcal{L}^{(i)}(\boldsymbol{\Lambda}) = \mathcal{L}^{(i)}_{\text{BR}}(\boldsymbol{\Lambda}) + \lambda \cdot \mathcal{KL}\left[\boldsymbol{t}^{(i)}\parallel \boldsymbol{\pi}^{(i)}\right]
\end{equation}
with coefficient $\lambda$ set to 0.5.

\subsection{Predictive distribution and confidence estimation}
Given $\operatorname{Dir}(\boldsymbol{\pi}^{(i)} | \boldsymbol{\alpha}^{(i)})$, the estimated probability of class $k$ can be calculated by the expectation of the predicted Dirichlet prior. Since the Dirichlet distribution is the conjugate prior of the categorical distribution, the expectation is tractable:
\begin{equation}
    \hat{\pi}_k^{(i)}=\mathbb{E}[{\pi}^{(i)}_k]=\frac{{\alpha}^{(i)}_k}{{\alpha_0}^{(i)}}
\end{equation}
where ${\alpha}_0^{(i)}= \sum_{k=1}^K \alpha_k^{(i)}$. Confidence is computed as follows:
\begin{gather}
 \hat{k}^{(i)}=\operatorname{argmax}_k \boldsymbol{\hat{\pi}}^{(i)}\\
    p^{(i)} = \hat{\pi}^{(i)}_{\hat{k}^{(i)}}
    \label{eqn: conf}
\end{gather}
where $\boldsymbol{\hat{\pi}}^{(i)}$ is the predictive distribution, $\hat{k}^{(i)}$ is the predicted class, and $p^{(i)}$ is the prediction confidence for a given input $\boldsymbol{x}^{(i)}$. Confidence is expected to reflect the probability of the output being correct.

\section{Evaluation metrics}
\label{sec: metric}
A range of metrics are adopted to evaluate the proposed method in terms of classification performance and model calibration.
The classification performance is evaluated by the accuracy (Acc) and F1 score (F1). Model calibration is evaluated by the expected calibration error (ECE), the normalised cross entropy (NCE), the area under the ROC curve (AUROC), and the area under the precision-recall curve (AUPRC).

Expected calibration error (ECE)~\cite{naeini2015obtaining} measures model calibration by computing the difference in expectation between confidence and accuracy. Confidence score is expected to reflect the probability of the predictions being actually correct. Predictions are partitioned into $Q$ bins equally spaced in the [0,1] range and ECE can be computed as follows:
\begin{equation}
    \text{ECE} = \sum_{q=1}^Q \frac{|B_q|}{n}\left| \text{Acc}(B_q) - \text{Conf}(B_q)\right|.
\end{equation}
A smaller ECE value indicates better model calibration.

Normalised cross entropy (NCE)~\cite{siu1997improved} measures how close the confidence score is to the probability of the predicted class being correct. Confidence scores for all test samples $p = [p_1, \ldots , p_N ]$ where $p_n \in [0, 1]$ are gathered and their corresponding target confidence $c = [c_1, \ldots , c_N ]$ where
$c_n \in \{0, 1\}$. NCE is then given by
\begin{equation}
    \operatorname{NCE}(\boldsymbol{c}, \boldsymbol{p})=\frac{\mathcal{H}(\boldsymbol{c})-\mathcal{H}(\boldsymbol{c}, \boldsymbol{p})}{\mathcal{H}(\boldsymbol{c})}
\end{equation}
where $\mathcal{H}(\boldsymbol{c})$ is the entropy of the target confidence sequence and $-\mathcal{H}(\boldsymbol{c}, \boldsymbol{p})$ is the binary cross-entropy between the target and the estimated confidence scores. When confidence estimation is systematically better than the correct ratio ($\sum_{n=1}^N c_n / N$), NCE is positive. For perfect confidence scores, NCE is 1.

AUROC is calculated as the area under the ROC curve. An ROC curve gives the trade-off between a true positive rate and a false positive rate across different decision thresholds. 
Similarly, AUPRC is calculated as the area under the precision-recall curve which shows the trade-off between precision and recall. 
The predicted confidence is used as the decision threshold for both AUROC and AUPRC. The baseline is 50\% for AUROC and is the fraction of positives for AUPRC.

\section{Experimental setup}
\label{sec: exp setup}
\subsection{Datasets}
The ADReSS dataset~\cite{luz20_interspeech} was used in this paper for automatic AD detection. It contains recordings of picture descriptions produced by cognitively normal subjects and patients with AD. The participants were asked to describe the ``cookie theft'' picture from the Boston diagnostic aphasia examination. The ADReSS dataset contains 156 participants, among which 78 (50\%) have been diagnosed with AD. The standard split of train/test provided by the corpus is used. 20\% of the training data was further set aside for validation.

The DAIC-WOZ data~\cite{DAIC-WOZ} was used in this paper which is a benchmark dataset for automatic depression detection. It consists of 189 clinical interviews between a participant and an animated virtual interviewer controlled by a human interviewer. 56 out of the 189 (30\%) recordings have been diagnosed as depressed.  The standard split of train/dev/test provided by the corpus is used.

\begin{figure}[t]
    \centering
    \includegraphics[width=0.7\linewidth]{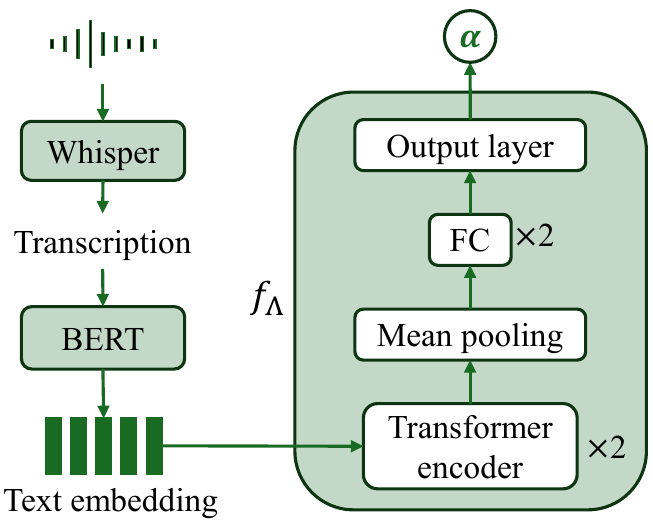}
    \caption{Illustration of the model structure. 
    }
    \label{fig: struc}
\end{figure}

\subsection{Model structure}
The model structure is shown in Figure~\ref{fig: struc}. Following~\cite{cui2023transferring}, the recording was first transcribed using a pretrained Whisper model\footnote{https://huggingface.co/openai/whisper-small}~\cite{radford2023robust}, which has a word error rate of 32.8\% on ADReSS and 20.4\% on DAIC-WOZ. The transcription was then encoded by a pretrained BERT model\footnote{https://huggingface.co/bert-base-uncased}~\cite{devlin-etal-2019-bert}. The model consists of two transformer encoder blocks~\cite{vaswani2017attention} of dimension 128 with four attention heads, followed by two fully connected (FC) layers with ReLU activation and an output layer with exponential activation to ensure $\boldsymbol{\alpha}$ is positive.

\subsection{Baselines}
\label{sec: baselines}
The proposed method is compared with the following baselines:
\begin{itemize}
    \item L2: a standard classification network with softmax output activation trained by cross-entropy loss with weight decay.
    \item MCDP: a Monte Carlo dropout~\cite{gal2016dropout} model with a dropout rate of 0.3 which is forwarded 50 times during testing with different dropout random seeds to obtain 50 samples.
    \item BBB: a Bayes by backprop~\cite{blundell2015weight} model which is forwarded 50 times during testing with different network weights to obtain 50 samples.
    \item Ensemble: an ensemble of five L2 models initialised and trained using different random seeds.
\end{itemize}
All baselines have the same backbone structure as the proposed method. Confidence is computed by Equation~\ref{eqn: conf}.

\begin{table}[t]
    \centering
    \caption{Comparison to the baselines in terms of classification accuracy and F1 score. Average and standard errors of five runs are reported. The best value in each column is shown in bold and the second best is underlined.}
    \label{tab:mean}
    \begin{tabular}{ccc}
        \toprule
 \multicolumn{3}{c}{\textbf{AD detection}}\\
 &  \textbf{F1} & \textbf{Acc} \\
\midrule
 L2&  {0.791$\pm$0.010}& 0.783$\pm$0.015\\
 MCDP&  0.786$\pm$0.010& 0.779$\pm$0.014\\
 BBB&  0.738$\pm$0.008& 0.721$\pm$0.025\\
 Ensemble&  \underline{0.792$\pm$0.011}& \underline{0.788$\pm$0.013}
\\
 Ours&  \textbf{0.807$\pm$0.013}& \textbf{0.800$\pm$0.019}\\
\midrule
\midrule
\multicolumn{3}{c}{\textbf{Depression detection}}\\
&  \textbf{F1 }& \textbf{Acc} \\
\midrule
 L2&  0.585$\pm$0.014& 0.706$\pm$0.018\\
 MCDP&  0.572$\pm$0.006& 0.695$\pm$0.034\\
 BBB&  0.580$\pm$0.012& 0.715$\pm$0.024\\
 Ensemble&  \textbf{0.602$\pm$0.008}& \underline{0.738$\pm$0.004}\\
 Ours&  \underline{0.600$\pm$0.008}& \textbf{0.745$\pm$0.008}\\
 \bottomrule
    \end{tabular}
    
\end{table}
\subsection{Implementation details}
The model was implemented using PyTorch. The AdamW optimiser and Noam learning rate scheduler were used with 400 warm-up steps and a peak learning rate of 4.29$\times 10^{-5}$.
Sub-dialogue shuffling~\cite{wu2023self} was used to augment and balance the training set which samples sub-dialogues $\boldsymbol{x}_{s:e}$ from each complete dialogue $\boldsymbol{x}_{1:T}$, where $s$ and $e$ are the randomly selected start and end sentence indices. The number of sub-dialogues for positive samples was set to 100 for AD and 500 for depression. 
Piece-wise linear mappings (PWLMs)~\cite{evermann2000large} were estimated on the validation set and then applied to the test set to better calibrate the confidence scores with accuracy. 
All experiments were run for 5 different seeds and the mean and standard error are reported.

\section{Results}
\label{sec: results}
In this section, the proposed method is compared to the baselines described in Section~\ref{sec: baselines} in terms of classification accuracy and confidence estimation.

\subsection{Classification performance}
Table~\ref{tab:mean} lists the classification accuracy and F1 scores for all compared methods. Comparing the proposed method to the baselines, it is shown that introducing a confidence measure does not degrade classification performance. The proposed method yields the best F1 and accuracy for AD detection as well as the highest accuracy for depression detection. Although the ensemble achieves the best prediction F1 score for depression detection, it involves training five individual systems. The proposed method achieves the second-best accuracy with only a fifth of the computational cost of Ensemble during training. 
During testing, the Ensemble involves five individual forward passes of each base model. Both MCDP and BBB involve 50 forward passes to obtain 50 samples. In contrast, the proposed method only requires a single forward pass and is thus the most efficient during testing.

\begin{table}[tb]
    \centering
    \caption{Comparison to the baselines in terms of ECE. A smaller ECE value indicates better model calibration.}
    \vspace{-1ex}
    \label{tab:ECE}
    \begin{tabular}{ccc}
    \toprule
         \textbf{AD}&  \textbf{ECE (w/o PWLM)}&  \textbf{ECE (w/ PWLM}) 
\\
\midrule
         L2&  0.227$\pm$0.016&  0.204$\pm$0.014
\\
         MCDP&  0.229$\pm$0.022&  0.207$\pm$0.005
\\
         BBB&  0.216$\pm$0.025&  0.195$\pm$0.013
\\
         Ensemble&  {0.173$\pm$0.012}&  0.153$\pm$0.017
\\
         Ours&  \textbf{0.163$\pm$0.011}&  \textbf{0.137$\pm$0.004}\\
         \midrule
         \midrule
         \textbf{Depression}& \textbf{ECE (w/o PWLM)}&  \textbf{ECE (w/ PWLM)} \\
         \midrule
L2&  0.312$\pm$0.023
&  0.217$\pm$0.008\\
         MCDP&  0.302$\pm$0.026
&  0.252$\pm$0.009\\
         BBB&  0.287$\pm$0.020
&  0.208$\pm$0.012\\
         Ensemble&  0.370$\pm$0.007&  0.219$\pm$0.008\\
         Ours&  \textbf{0.207$\pm$0.009}&  \textbf{0.183$\pm$0.009}\\
         \bottomrule
    \end{tabular}
    
\end{table}
\begin{table}[tb]
    \centering
    \caption{Comparison to the baselines in terms of NCE. A larger NCE value indicates better confidence estimation.}
    \vspace{-1ex}
    \label{tab:NCE}
    \begin{tabular}{ccc}
    \toprule
     \textbf{AD}     &  \textbf{NCE (w/o PWLM)} &\textbf{NCE (w/ PWLM)} \\
         \midrule
         L2
&  -0.138$\pm$0.047&0.103$\pm$0.011
\\
         MCDP
&  -0.083$\pm$0.032&0.121$\pm$0.010
\\
         BBB
&  -0.140$\pm$0.63&0.095$\pm$0.028
\\
         Ensemble
&  0.028$\pm$0.017&0.141$\pm$0.035
\\
         Ours&  \textbf{0.066$\pm$0.038}&\textbf{0.210$\pm$0.030} \\
         \midrule
         \midrule
\textbf{Depression} & \textbf{NCE (w/o PWLM)} &\textbf{NCE (w/ PWLM)}\\
\midrule
         L2
&  -0.288$\pm$0.079&0.046$\pm$0.012
\\
         MCDP
&  -0.116$\pm$0.052&0.100$\pm$0.011
\\
         BBB
&  -0.171$\pm$0.077&0.073$\pm$0.017
\\
 Ensemble
& -0.307$\pm$0.079&0.045$\pm$0.010
\\
 Ours& \textbf{0.048$\pm$0.028}&\textbf{0.155$\pm$0.007} \\
 \bottomrule
    \end{tabular}
    
\end{table}

\subsection{Confidence estimation}
The ECE values and NCE values before and after applying PWLM are shown in Table~\ref{tab:ECE} and Table~\ref{tab:NCE} respectively. 
PWLM improves both ECE and NCE for most methods while it has a larger impact on NCE than ECE. The proposed method performs the best both before and after applying PWLM. %

\begin{figure}[tb]
    \centering
    \begin{minipage}[b]{0.45\linewidth}
    \centerline{\includegraphics[width=\linewidth]{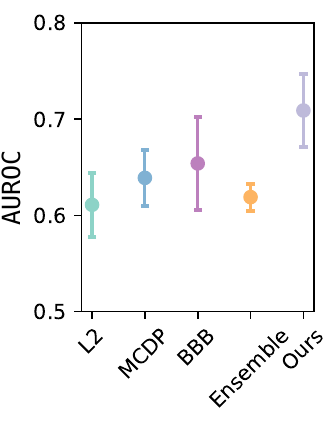}}
      \centerline{(a) AUROC}
    \end{minipage}
    \begin{minipage}[b]{0.45\linewidth}
    \centerline{\includegraphics[width=\linewidth]{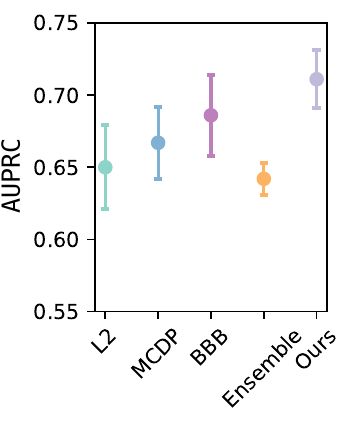}}
      \centerline{(b) AUPRC}
    \end{minipage}    
    \vspace{-1ex}
    \caption{Comparison to the baselines in terms of AUROC and AUPRC for AD detection. The average of five runs is plotted along with standard error as error bars.}
    \label{fig:AD-AUC}
\end{figure}

\begin{figure}[tb]
    \centering
    \begin{minipage}[b]{0.45\linewidth}
    \centerline{\includegraphics[width=\linewidth]{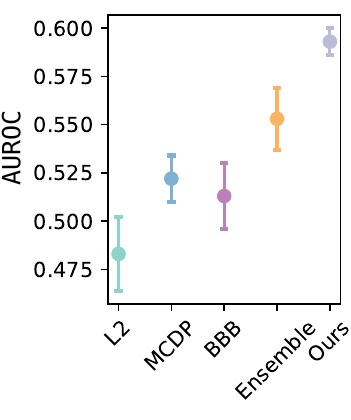}}
      \centerline{(a) AUROC}
    \end{minipage}
    \begin{minipage}[b]{0.45\linewidth}
    \centerline{\includegraphics[width=\linewidth]{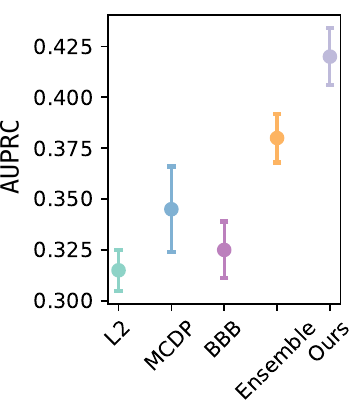}}
      \centerline{(b) AUPRC}
    \end{minipage}    
    \vspace{-1ex}
    \caption{Comparison to the baselines in terms of AUROC and AUPRC for depression detection. }
    \label{fig:Dep-AUC}
    \vspace{-1ex}
\end{figure}

Since PWLM is monotonic, NCE and ECE values will be affected while AUROC and AUPRC remain unchanged as the relative order of confidence scores is unchanged. The AUROC and AUPRC are compared in Figure~\ref{fig:AD-AUC} and Figure~\ref{fig:Dep-AUC} for the detection of AD and depression respectively. The proposed method performs the best in terms of both AUROC and AUPRC on both datasets, which further demonstrates its superior capability of confidence estimation\footnote{The proposed method consistently outperforms the baselines in terms of all confidence estimation metrics across all five seeds.}.

\subsection{Analysis: reject option}
A reject option was added 
where the diagnosis system has the option to accept or decline a sample based on 
the confidence. 
From Table~\ref{tab:threshold}, improved classification performance is observed for both AD and depression detection when the confidence threshold is increased from 0.5 to 0.8, \textit{i.e.},  
the model provides more accurate predictions when it is more confident, which indicates the effectiveness of the proposed confidence estimation. 
\begin{table}[H]
    \centering
    \caption{A reject option based on confidence score.}
    \vspace{-1ex}
    \label{tab:threshold}
    \begin{tabular}{ccccc}
    \toprule
         \textbf{Confidence} &  \multicolumn{2}{c}{\textbf{AD}}&  \multicolumn{2}{c}{\textbf{Depression}}\\
         
 \textbf{threshold}& \textbf{F1}& \textbf{Acc}& \textbf{F1}&\textbf{Acc}\\
 \midrule
         50\%&  0.807& 0.800& 0.600& 0.745\\
         80\%&  0.913&  0.920&  0.755& 0.831\\
         \bottomrule
    \end{tabular}
\end{table}

\section{Conclusions}
This paper investigates confidence estimation of automatic detection of Alzheimer's disease and depression. A novel Bayesian approach is proposed which places a dynamic Dirichlet prior over the categorical likelihood to model the second-order uncertainty of model prediction. A range of metrics are adopted to evaluate the performance for detection accuracy and confidence estimation. Experiments were conducted on the AD dataset ADReSS and depression dataset DAIC-WOZ. Results show that the proposed method clearly and consistently outperforms a range of baselines in terms of both classification accuracy and confidence estimation for both Alzheimer's disease detection and depression detection. It is hoped that the proposed method could help promote reliable and trustworthy automatic diagnostic systems. Although this work focuses on detection based on speech information from clinical interviews, the proposed method should also work for other input modalities (\textit{i.e.} image) and the confidence estimation could also be useful when combining different systems.

\bibliographystyle{IEEEtran}
\bibliography{mybib}

\end{document}